\documentclass[10pt,twocolumn,letterpaper]{article}

\usepackage{cvpr}
\usepackage{times}
\usepackage{epsfig}
\usepackage{graphicx}
\usepackage{amsmath}
\usepackage{amssymb}
\usepackage{enumitem}
\usepackage[ruled,linesnumbered]{algorithm2e}

\usepackage[pagebackref=true,breaklinks=true,letterpaper=true,colorlinks,bookmarks=false]{hyperref}

 \cvprfinalcopy 

\def\cvprPaperID{495} 

\ifcvprfinal\fi
\begin{document}

\title{DeepID-Net: Deformable Deep Convolutional Neural Networks \\for Object Detection }

\author{ Wanli Ouyang,  Xiaogang Wang, Xingyu Zeng, Shi Qiu, Ping Luo, Yonglong Tian, \\
 Hongsheng Li, Shuo Yang, Zhe Wang,  Chen-Change Loy, Xiaoou Tang \\
The Chinese University of Hong Kong\\
{\tt\small {wlouyang, xgwang}@ee.cuhk.edu.hk}
}

\maketitle

\begin{abstract}
In this paper, we propose deformable deep convolutional neural networks for generic object detection. This new deep learning object detection framework has innovations in multiple aspects. In the proposed new deep architecture, a new deformation constrained pooling (def-pooling) layer models the deformation of object parts with geometric constraint and penalty. A new pre-training strategy is proposed to learn feature representations more suitable for the object detection task and with good generalization capability. By changing the net structures, training strategies, adding and removing some key components in the detection pipeline, a set of models with large diversity are obtained, which significantly improves the effectiveness of model averaging. The proposed approach improves the mean averaged precision obtained by RCNN \cite{girshick2014rich}, which was the state-of-the-art, from $31\%$ to $50.3\%$ on the ILSVRC2014 detection test set. It also outperforms the winner of ILSVRC2014, GoogLeNet, by 6.1\%. Detailed component-wise analysis is also provided through extensive experimental evaluation, which provide a global view for people to understand the deep learning object detection pipeline. 
\end{abstract}

\section{Introduction}
\label{Sec:Introduction}
Object detection is one of the fundamental challenges in computer vision.  It has attracted a great deal of research interest \cite{Dalal:HOG,Smeulders:SelectiveSearch,LatSVMObj, he2014spatial}. Intra-class variation in appearance and deformation are among the main challenges of this task. 

\begin{figure}
\centering
\centerline{\epsfig{figure=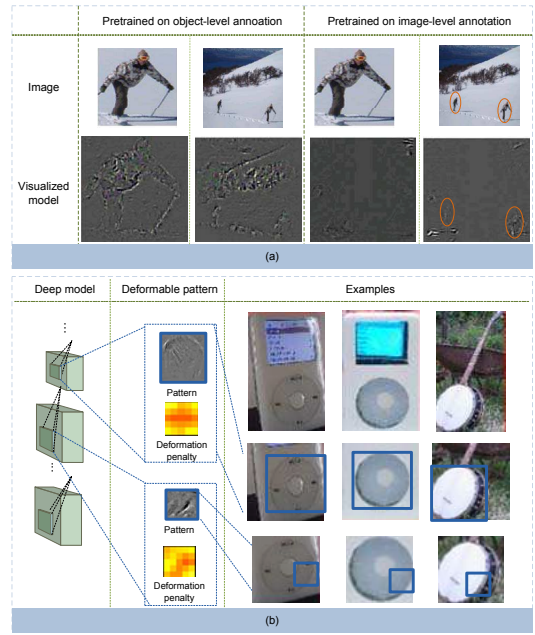,width=7cm}}
\caption{The motivation of this paper in new pretraining scheme (a) and jointly learning feature representation and deformable object parts shared by multiple object classes at different semantic levels (b). In (a), a model pretrained on image-level annotation is more robust to size and location change while a model pretrained on object-level annotation is better in representing objects with tight bounding boxes. In (b), when ipod rotates, its circular pattern moves horizontally at the bottom of the bounding box. Therefore, the circular patterns have smaller penalty moving horizontally but higher penalty moving vertically. The curvature part of the circular pattern are often at the bottom right positions of the circular pattern. \emph{Best viewed in color}.}
\label{Fig:Observation}
\vspace{-10pt}
\end{figure}

Because of its power in learning features, the convolutional neural network (CNN) is being widely used in  recent large-scale object detection and recognition systems \cite{szegedy2014going, simonyan2014very, he2014spatial, lin2013network}.
Since training deep models is a non-convex optimization problem with millions of parameters, the choice of a good initial point is a crucial but unsolved problem, especially when deep CNN goes deeper \cite{szegedy2014going, simonyan2014very, lin2013network}.
It is also easy to overfit to a small training set. Researchers  find that supervised pretraining on large-scale image classification data and then finetuning for the targeting object detection task is a practical solution \cite{donahue2014decaf, razavian2014cnn, zhang2014part, girshick2014rich}. However, we observe that there is still a gap between the pretraining task and the finetuning task that makes pretraining
less effective. The problem of the training scheme  is the mismatch between pretraining with the image classification task and fine-tuning for the object detection task.
For image classification, the input is a whole image and the task is to recognize the object within this image.
Therefore,  learned feature representations  have robustness to  scale and location change of objects in images.
Taking Fig. \ref{Fig:Observation}(a) as an example, no matter how large and where a person is in the image, the image should be classified as person. However, robustness to object size and location is not required for object detection. For object detection, candidate regions are cropped and warped before they are used as input of the deep model. Therefore, the positive candidate regions for the object class person should have their locations aligned and their sizes normalized. On the contrary, the deep model is expected to be sensitive to the change on position and size in order to accurately localize objects. An example to illustrate the mismatch is shown in Fig. \ref{Fig:Observation} (a). Because of such mismatch, the image classification task is not an ideal choice to pretrain the deep model for object detection. Therefore, a new pretraining scheme is proposed to train the deep model for object detection more effectively.


Part deformation handling is a key factor for the recent progress in generic object detection \cite{LatSVMObj, Zhulong:StructSVM,PictorialStruct, yang2013articulated}. Our new CNN layer is motivated by three observations. First, deformable visual patterns are shared by objects of different categories. For example, the circular visual pattern is shared by both banjo and ipod as shown in Fig. \ref{Fig:Observation}(b). Second,  the regularity on deformation exists for visual patterns at different semantic levels. For example, human upper bodies, human heads, and human mouths are parts at different semantic levels with different deformation properties. Third, a deformable part at a higher level is composed of  deformable parts at a lower level. For example, a human upper body is composed of a head and other body parts. With these observations, we design a new deformation-constrained pooling (def-pooling) layer to learn the shared visual patterns and their deformation properties for multiple object classes at different semantic levels and composition levels.

The performance of deep learning object detection systems depends significantly on implementation
details \cite{chatfield2014return}.
However, an evaluation of the performance of the recent deep architectures on the common ground for large-scale object detection is missing. As a respect to the devil of details \cite{chatfield2014return, girshick2014rich}, this paper compares the performance of recent deep models, including AlexNet \cite{Krizhevsky:ImageNetCNN}, ZF \cite{zeiler2013visualizing}, Overfeat \cite{sermanet2013overfeat}, and GoogleNet \cite{szegedy2014going} under the same setting for different pretraining-finetuining schemes.

%

In this paper, we propose a deformable deep convolutional neural network for object detection; named as DeepID-Net. In DeepID-Net, we jointly learn the feature representation and part deformation for a large number of object categories. We also investigate many aspects in effectively and efficiently training and aggregating the deep models, including bounding box rejection, training schemes, context modeling, and model averaging. The proposed new framework significantly advances the state-of-the-art for deep learning based generic object detection, such as the well known RCNN \cite{girshick2014rich} framework. 
This paper also provides detailed component-wise experimental results on how our approach can improve the mean Averaged Precision (AP) obtained by RCNN \cite{girshick2014rich} from 31.0\% to mean AP 50.3\% step-by-step on the ImageNet Large Scale Visual Recognition Challenge 2014 (ILSVRC2014) object detection task.

The contributions of this paper are as follows:
\begin{enumerate}[leftmargin=12pt,noitemsep,nolistsep]
\item A new deep learning framework for object detection. It effectively integrates feature representation learning, part deformation learning, context modeling, model averaging, and bounding box location refinement into the detection system. Detailed component-wise analysis is provided through extensive experimental evaluation. This paper is also the first to investigate the influence of CNN structures for the large-scale object detection task under the same setting. By changing the configuration of this framework, multiple detectors with large diversity are generated, which leads to more effective model averaging.
\item A new scheme for pretraining the deep CNN model. We propose to pretrain the deep model on the ImageNet image classification and localization dataset with 1000-class object-level annotations instead of with image-level annotations, which are commonly used in existing deep learning object detection \cite{girshick2014rich, szegedy2014going}. Then the deep model is fine-tuned on the ImageNet/PASCAL-VOC object detection dataset with 200/20 classes, which are the targeting object classes in the two datasets. 
\item A new deformation constrained pooling (def-pooling) layer, which enriches the deep model by learning the deformation of object parts at any information abstraction levels. The def-pooling layer can be used for replacing the max-pooling layer and learning the deformation properties of parts. 
\end{enumerate}

\section{Related work}

Since many objects have non-rigid deformation, the ability to handle deformation improves detection performance. Deformable part-based models were used in \cite{LatSVMObj, Zhulong:StructSVM} for handling translational movement of parts. To handle more complex articulations, size change and rotation of parts were modeled in \cite{PictorialStruct}, and mixture of part appearance and articulation types were modeled in \cite{Bourdev:Poslet, Yang:Articulated}. A dictionary of shared deformable patterns was learned in  \cite{Hariharan2014Detecting}. In these approaches, features are manually designed.

Because of the power on learning feature representation, deep models have been widely used for object recognition, detection and other vision tasks \cite{sermanet2013overfeat, zeiler2013visualizing,he2014spatial,Simonyan14a,zou2014generic, gong2014multi, lin2013network, girshick2014rich, Ouyang:DBNHuman, ouyang2013DBN2Ped,Zeng2014Deep,Zeng2013Multi,ouyang2014multi,Luo2014Switchable,Sun2013Deep,sun2014deep,Sun2013Deep,Sun2013Hybrid,Luo2013Deep,Luo2013Pedestrian,Luo:DeepFace, zhao2015saliency}. In existing deep CNN models, max pooling and average pooling are useful in handling deformation but cannot learn the deformation penalty and geometric models of object parts. The deformation layer was first proposed in \cite{Ouyang2013JointDeep} for pedestrian detection. In this paper, we extend it to general object detection on ImageNet. In \cite{Ouyang2013JointDeep}, the deformation layer was constrained to be placed after the last convolutional layer, while in this work the def-pooling layer can be placed after all the convolutional layers to capture geometric deformation at all the information abstraction levels. In \cite{Ouyang2013JointDeep}, it was assumed that a pedestrian only has one instance of a body part, so each part filter only has one optimal response in a detection window. In this work, it is assumed that an object has multiple instances of a part (e.g. a car has many wheels), so each part filter is allowed to have multiple response peaks in a detection window. Moreover, we allow multiple object categories to share deformable parts and jointly learn them with a single network. This new model is more suitable for general object detection. 

Context gains attentions in object detection.
The context information investigated in literature  includes regions surrounding objects \cite{Dalal:HOG, Ding:ContextPedDet,galleguillos2010multi}, object-scene interaction \cite{Divvala:ContextObjDet, heitz2008learning}, and the presence, location, orientation and size relationship among objects \cite{Barinova:DetHough, Wu:IJCV07,Yan:PedDet,Desai:ObjLayout, Park:MultResObjDet, galleguillos2010multi, Song:ObjDetContext,Divvala:ContextObjDet,Yao:posObj,Ding:ContextPedDet,Yang:Proxemics,ouyang2013DBN2Ped, Desai:PhraseletsECCV12, sadeghi2011recognition, tang2013learning}. 
In this paper, we use whole-image classification scores over a large number of classes from a deep model as  global contextual information to refine detection scores.


Besides feature learning, deformation modeling, and context modeling, there are also other important components in the object detection pipeline, such as pretraining \cite{girshick2014rich}, network structures \cite{sermanet2013overfeat, zeiler2013visualizing,  Krizhevsky:ImageNetCNN}, refinement of bounding box locations \cite{girshick2014rich}, and model averaging \cite{zeiler2013visualizing, Krizhevsky:ImageNetCNN, he2014spatial}. While these components were studies individually in different works, we integrate them into a complete pipeline and take a global view of them with component-wise analysis under the same experimental setting. It is an important step to understand and advance deep learning based object detection.

\section{Method}

\subsection{Overview of our approach}
\label{ssec:overview}
\begin{figure} 
\centering
\centerline{\epsfig{figure=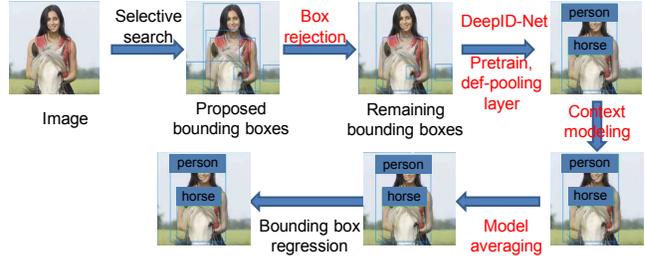,width=8.5cm}}
\caption{Overview of our approach. 
Find detailed description in the text of Section \ref{ssec:overview}.
Texts in red highlight the steps that are not present in RCNN \cite{girshick2014rich}.}
\label{Fig:ModelAll}
\vspace{-10pt}
\end{figure}

An overview of our proposed approach is shown in Fig. \ref{Fig:ModelAll}. 
We take the ImageNet object detection task as an example. The ImageNet image classification and localization dataset with 1,000 classes is chosen to pretrain the deep model. Its object detection dataset has 200 object classes. In the experimental section, the approach is also applied to the PASCAL VOC. The pretraining data keeps the same, while the detection dataset only has 20 object classes. The steps of our approach are summarized as follows.
\begin{enumerate}[leftmargin=12pt,noitemsep,nolistsep]
\item  Selective search proposed in \cite{Smeulders:SelectiveSearch} is used to propose candidate bounding boxes. 
\item An existing detector, RCNN \cite{girshick2014rich} in our experiment,  is used to reject bounding boxes that are most likely to be background.
\item An image region in a bounding box is cropped and fed into the DeepID-Net to obtain 200 detection scores. Each detection score measures the confidence on the cropped image containing one specific object class. Details are given in Section \ref{Sec:DeepIdNet}.
\item The 1000-class whole-image classification scores of a deep model are used as  contextual information to refine the detection scores of each candidate bounding box. Details are given in Section \ref{Sec:Context}.
\item Average of multiple deep model outputs is used to improve the detection accuracy. Details are given in Section \ref{Sec:CombineModel}. 
\item Bounding box regression proposed in RCNN \cite{girshick2014rich} is used to reduce localization errors. 
\end{enumerate}

\subsection{Architecture  of DeepID-Net}
\label{Sec:DeepIdNet}
\label{Sec:NetOverview}
\begin{figure} 
\centering
\begin{tabular}{c}
\centerline{\epsfig{figure=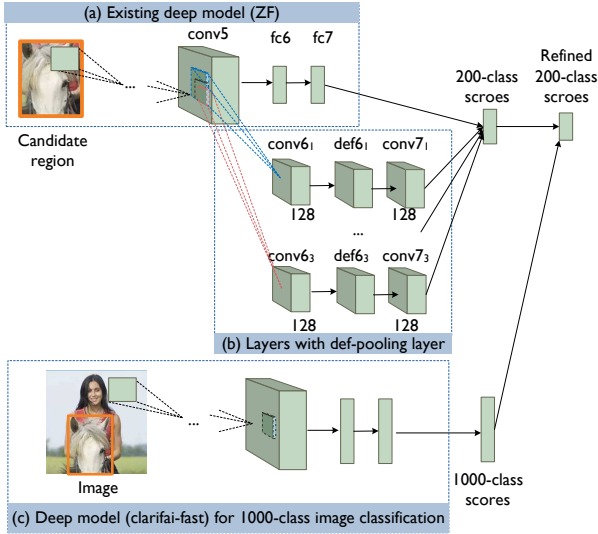,width=8cm}} \\
\end{tabular}
\vspace{-5pt}
\caption{Architecture of DeepID-Net with three parts: (a) baseline deep model, which is  ZF \cite{zeiler2013visualizing} in our best-performing single-model detector; (b)  layers of part filters with variable sizes and def-pooling layers; (c)  deep model to obtain 1000-class image classification scores. The 1000-class image classification scores are used to refine the 200-class bounding box classification scores.}
\label{Fig:DeepIDmodel}
\vspace{-15pt}
\end{figure}

DeepID-Net in Fig. \ref{Fig:DeepIDmodel} has three parts:
\begin{enumerate}[label=(\alph*),leftmargin=12pt,noitemsep,nolistsep]
\item The baseline deep model. The ZF model proposed in \cite{zeiler2013visualizing} is used as the default baseline deep model when it is not specified. 
\item Branches with def-pooling layers. The input of these layers is the conv5, the last convolutional layer of the baseline model. The output of conv5 is convolved with part filters of variable sizes and the proposed def-pooling layers in Section \ref{Sec:DeformLayer} are used to learn the deformation constraint of these part filters. Parts (a)-(b) output 200-class object detection scores. For the cropped image region that contains a horse as shown in Fig. \ref{Fig:DeepIDmodel}(a), its ideal output should have a high score for the object class horse but low scores for other classes.
\item The deep model (ZF) to obtain image classification scores of 1000 classes. Its input is the whole image, as shown in Fig. \ref{Fig:DeepIDmodel}(c). The image classification scores are used as contextual information to refine the classification scores of  bounding boxes. Detail are given in Section \ref{Sec:Context}.
\end{enumerate}

\subsection{New pretraining strategy}
\label{Sec:Prtrain}
The widely used training scheme in deep learning based object detection \cite{girshick2014rich, zhang2014part, szegedy2014going} including RCNN is denoted by Scheme 0 and described as follows:
\begin{enumerate}[leftmargin=12pt,noitemsep,nolistsep]
\item Pretrain  deep models by using the image classification task, i.e. using image-level annotations from the ImageNet image classification and localization training data.
\item Fine-tune  deep models for the object detection task, i.e. using object-level annotations from the object detection training data. The parameters learned in Step (1) are used  as initialization.
\end{enumerate}
The deep model structures at the pretraining and fine-tuning stages are only different in the last fully connected layer for predicting labels ($1,000$ classes for the ImageNet classification task vs. $200$ classes for the ImageNet detection task). Except for the last fully connected layers for classification, the parameters learned at the pretraining stage are directly used as initial values for the fine-tuning stage.


We propose to pretrain the deep model on a large auxiliary object detection training data instead of the image classification data. Since the ImageNet Cls-Loc data provides object-level bounding boxes for 1000 classes, more diverse in content than the ImageNet Det data with 200 classes, we use the image regions cropped by these bounding boxes to pretain the baseline deep model in Fig. \ref{Fig:DeepIDmodel}(a). The proposed  pretraining strategy is denoted as Scheme 1 and  bridges the image- vs. object-level annotation gap in RCNN. 



\begin{enumerate}[leftmargin=12pt,noitemsep,nolistsep]
\item Pretrain the deep model with object-level annotations of $1,000$ classes from ImageNet Cls-Loc train data.
\item Fine-tune the deep model for the 200-class object detection task, i.e. using object-level annotations of 200 classes from ImageNet Det train and val$_1$ (validation set 1) data.  Use the parameters in Step (1) as initialization.
\end{enumerate}
Compared with the training scheme of RCNN, experimental results show that the proposed scheme improves mean AP by 4.5\%  on ImageNet Det val$_2$ (validation set 2). If only the 200 targeting classes (instead of 1,000 classes) from the ImageNet Cls-Loc train data are selected for pre-training in Step (1), the mean AP on ImageNet Det val$_2$ drops by 5.7\%.





\subsection{Def-pooling layer}
\label{Sec:DeformLayer}
In the deformable part based model (DPM) \cite{LatSVMObj}  for object detection,  part templates learned on HOG features are considered as part filters and they are convolved with input images. Similarly, we can consider the input of a convolutional layer in CNN as features and consider the filters of that convolutional layer as part filters. And the outputs of the convolutional layer are part detection maps. 

Similar to max-pooling and average-pooling, the input of a def-pooling layer is the output of a convolutional layer.  The convolutional layer produces $C$ part detection maps of size $W\times H$.  
Denote $\mathbf{M}_c$ as the $c$th part detection map. Denote the $(i,j)$th element of $\mathbf{M}_c$ by $m_c^{(i,j)}$.
The def-pooling layer takes a small block with center $(s_x\cdot x, s_y\cdot y)$ and size $(2R+1)\times (2R+1)$ from the $\mathbf{M}_c$ and produce the element of the output as follows: 
{\small
\vspace{-5pt}
\begin{equation}
\label{eq:GenDefMap2}
\begin{split}
b_c^{(x,y)} &= \max_{\delta_{x},\delta_{y}\in\{-R, \cdots, R\}}{\{m^{\mathbf{z}_{\delta_{x},\delta_{y}} }_c - \sum_{n=1}^N a_{c,n}d^{\delta_{x},\delta_{y}}_{c, n}\}},  \\
&\textrm{where }
 \mathbf{z}_{\delta_{x},\delta_{y}} = (s_x\cdot x+\delta_{x}, s_y\cdot y+\delta_{y}).
\end{split}
\vspace{-5pt}
\end{equation}}


\begin{itemize}[leftmargin=12pt,noitemsep,nolistsep]
\item $b_c^{(x,y)}$ is the $(x,y)$th element of the output of the def-pooling layer. For $\mathbf{M}_c$ of size $W \times H$, the subsampled output has size $\frac{W}{s_x}\times \frac{H}{s_y}$. Therefore, multiple max responses are allowed for each part filer. 
\item $m^{\mathbf{z}_{\delta_{x},\delta_{y}} }_c$ is the visual score of placing the $c$th part at the deformed position $\mathbf{z}_{\delta_{x},\delta_{y}}$.
\item  $a_{c, n}$ and $d^{\delta_{x},\delta_{y}}_{c, n}$ are parameters learned by BP.  $\sum_{n=1}^N a_{c, n} d^{\delta_{x},\delta_{y}}_{c, n}$ is the penalty of placing the part from the assumed anchor position $ (s_x\cdot x, s_y\cdot y)$ to the deformed position $\mathbf{z}_{\delta_{x},\delta_{y}}$. 
\end{itemize}

The def-pooling layer can be better understood through the following examples.  


\emph{Example 1.}  If $N=1$, $a_n=1$, $d^{\delta_{x}, \delta_{y}}_{1}$ = 0 for $|\delta_{x}|, |\delta_{y}| \leq k$ and $d^{\delta_{x}, \delta_{y}}_{1}= \infty$ for $|\delta_{x}|, |\delta_{y}| > k$, then this corresponds to max-pooling with kernel size $k$.  It shows that the max-pooling layer is a special case of the def-pooling layer.  Penalty becomes very large when deformation reaches certain range. Since the use of different kernel sizes in max-pooling corresponds to different maps of deformation penalty that can be learned by BP, def-pooling provides the ability to learn the map that implicitly decides the kernel size for max-pooling.

\emph{Example 2.} The deformation layer in \cite{Ouyang2013JointDeep} is a special case of the def-pooling layer by enforcing that $\mathbf{z}_{\delta_{x}, \delta_{y}}$ in (\ref{eq:GenDefMap2}) covers all the locations in conv$_{l-1,i}$ and only one output with a pre-defined location is allowed for the def-pooling layer (i.e. $R= \infty$, $s_x = W$, and $s_y = H$). The proof can be found in Appendix 1. To implement quadratic deformation penalty used in \cite{LatSVMObj}, we can predefine $\{d^{\delta_{x},\delta_{y}}_{c, n}\}_{n=1,2,3,4}=\{\delta_{x}, \delta_{y}, (\delta_{x})^2, (\delta_{y})^2\}$ and learn parameters $a_n$. As shown in Appendix A, the def-pooling layer under this setting can represent deformation constraint in the deformable part based model (DPM) \cite{LatSVMObj} and the DP-DPM \cite{girshick2014deformable}.

\begin{figure} 
\centering
\centerline{\epsfig{figure=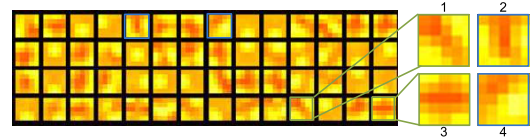,width=7cm}}
\caption{The learned deformation penalty for different visual patterns. The penalties in map 1 are low at diagonal positions. The penalties in map 2 and 3 are low at vertical and horizontal locations separately. The penalties in map 4 are high at the bottom right corner and low at the upper left corner. }
\label{Fig:DefLayerParam}
\end{figure}

\emph{Example 3.} If $N=1$ and $a_n=1$, then $d^{\delta_{x}, \delta_{y}}_{1}$ is the deformation parameter/penalty of moving a part from the assumed location $(s_x\cdot x,s_y\cdot y)$ by $(\delta_{x}, \delta_{y})$. If the part is not allowed to move, we have $d^{0,0}_{1}=0$ and $d^{\delta_{x}, \delta_{y}}_{1}=\infty$ for $(\delta_{x}, \delta_{y})\neq (0,0)$.  If the part has  penalty 1 when it is not at the assumed location $(s_x\cdot x,s_y\cdot y)$, then we have $d^{0,0}_{1}=0$  and $d^{\delta_{x}, \delta_{y}}_{1}=1$ for $(\delta_{x}, \delta_{y})\neq (0,0)$. It allows to assign different penalty to displacement in different directions. If the part has penalty 2 moving leftward and penalty 1 moving rightward, then we have $d^{\delta_{x}, \delta_{y}}_{1}=1$ for $\delta_{x}<0$ and $d^{\delta_{x}, \delta_{y}}_{1}=2$ for $\delta_{x}>0$.
Fig. \ref{Fig:DefLayerParam} shows some learned deformation parameters $d^{\delta_{x}, \delta_{y}}_{1}$.

Take Example 2 as an example for BP learning. $a_{c,n}$ is the parameter in this layer and $d_*$ is pre-defined constant. $\partial b_c^{(x,y)}/\partial a_{c,n}=-d_{c,n}^{(\delta_x, \delta_y)}$  for the position $(\delta_x, \delta_y)$ with maximum value in (\ref{eq:GenDefMap2}). The gradients for the parameters in the layers before the def-pooling layer are back-propagated like max-pooling layer. 

In our implementation, there are no fully connected layers after conv$7_1,2,3$ in Fig. \ref{Fig:DeepIDmodel} and Example 3 is used for def-pooling.

\begin{figure} 
\centering
\centerline{\epsfig{figure=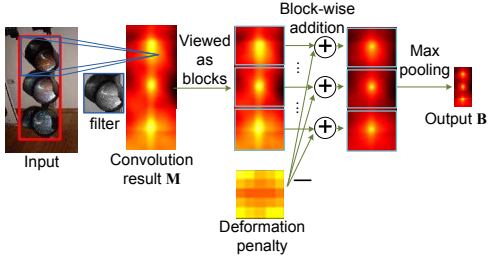,width=6.5cm}}
\caption{Def-pooling layer. The part detection map and the deformation penalty are summed up. Block-wise max pooling is then performed on the summed map to obtain the output $\mathbf{B}$ of size $\frac{H}{s_y}\times \frac{W}{s_x}$ ($3 \times 1$ in this example). }
\label{Fig:DefLayerGen}
\vspace{-20pt}
\end{figure}

\subsubsection{Analysis}
\begin{figure} 
\centering
\centerline{\epsfig{figure=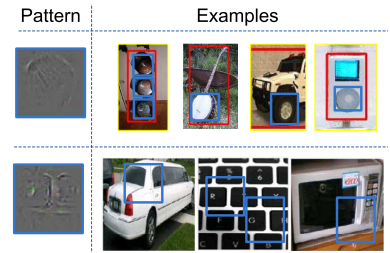,width=5cm}}
\caption{Repeated visual patterns in multiple object classes.}
\label{Fig:Share_pattern}
\vspace{-10pt}
\end{figure}

A visual pattern has different spatial distributions in different object classes.
For example, traffic lights and ipods have geometric constraints on the circular visual pattern in Fig. \ref{Fig:Share_pattern}. The weights connecting the convolutional layers conv7$_1$ - conv7$_3$ in Fig. \ref{Fig:DeepIDmodel} and classification scores are determined by the spatial distributions of visual patterns for different classes. For example, the car class will have large positive weights in the bottom region but negative weights in the upper region for the circular pattern. On the other hand, the traffic light class will have positive weights in the upper region for the circular pattern. 

A single output of the convolutional layer conv7$_1$ in Fig. \ref{Fig:DeepIDmodel} is from multiple part scores in def6$_1$. The relationship between parts of the same layer is modeled by conv7$_1 $.

The def-pooling layer has the following advantages. 

\begin{enumerate}[leftmargin=12pt,noitemsep,nolistsep]
\item 
It can replace any convolutional layer, and learn  deformation of parts with different sizes and semantic meanings. For example, at a higher level, visual patterns can be large parts, e.g. human upper bodies, and  the def-pooling layer can capture the deformation constraint of human upper parts. At a middle level,  the visual patterns can be smaller  parts, e.g. heads. At the lowest level, the visual patterns can be very small, e.g. mouths. A human upper part is composed of a deformable head and other parts. The human head is composed of a deformable mouth and other parts. Object parts at different semantic abstraction levels with different deformation constraints are captured by def-pooling layers at different levels. The composition of object parts is naturally implemented by CNN with  hierarchical layers.

\item 
The def-pooling layer allows for multiple deformable parts with the same visual cue, i.e. multiple response peaks are allowed for one filter. This design is from our observation that an object may have multiple object parts with the same visual pattern. For example, three light bulbs co-exist in a traffic light in Fig. \ref{Fig:DefLayerGen}.
 
\item 
As shown in Fig. \ref{Fig:DeepIDmodel}, the def-pooling layer is a shared representation for multiple  classes and therefore the learned visual patterns in the def-pooling layer can be shared among these classes.
As examples in Fig. \ref{Fig:Share_pattern}, the learned circular visual patterns are shared as different object parts in traffic lights, cars, and ipods. 
\end{enumerate}
The layers proposed in \cite{Ouyang2013JointDeep, girshick2014deformable}  does not have these advantages, because they can only be placed after the final convolutional layer, assume one instance per object part, and does not share visual patterns among classes.

\subsection{Fine-tuning the deep model with hinge-loss}

In RCNN, feature representation is first learned with the softmax loss in the deep model after fine-tuning. Then in a separate step, the learned feature representation is input to a linear binary SVM classifier for detection of each class. 
In our approach, the softmax loss is replaced by the 200 binary hinge losses when fine-tuning the deep model. Thus the deep model fine-tuning and SVM learning steps in RCNN are merged into one step. The extra training time  required for extracting features ($\sim$ 2.4 days with one Titan GPU) is saved. 

\subsection{Contextual modeling}
\label{Sec:Context}
The deep model learned for the image classification task (Fig. \ref{Fig:DeepIDmodel} (c)) takes  scene information into consideration while the deep model for object detection (Fig. \ref{Fig:DeepIDmodel} (a) and (b)) focuses on local bounding boxes. 
The 1000-class image classification scores are used as contextual features, and concatenated with the 200-class object detection scores to form a 1200 dimensional feature vector, based on which a linear SVM is learned to refine the 200-class detection scores.

\begin{figure} 
\centering
\centerline{\epsfig{figure=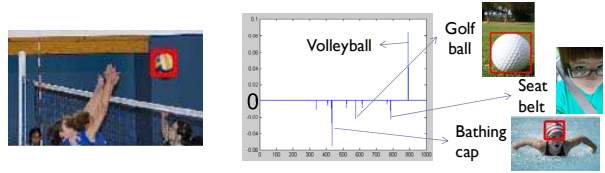,width=8cm}}
\caption{The SVM weights on image classification scores (a) for the object detection class volleyball (b). }
\vspace{-10pt}
\label{Fig:Context}
\end{figure}

Take object detection for class volleyball as an example in Figure \ref{Fig:Context}.  
If only considering local  regions cropped from bounding boxes, volleyballs are easy to be confused with bathing caps and golf balls. In this case, the contextual information from the whole-image classification scores is helpful, since bathing caps  appear in scenes of beach and swimming pools,  golf balls appear in grass fields, and volleyballs appear in stadiums.  The whole images of the three classes can be better distinguished because of the global scenery information. 
Fig. \ref{Fig:Context} plots the learned linear SVM weights on the 1000-class image classification scores. It is observed that image classes bathing cap and golf ball suppress the existence of volleyball in the refinement of detection scores with negative weights, while the image class volleyball enhances the detection score of volleyball. 

\subsection{Combining models with high diversity}
\label{Sec:CombineModel}
Model averaging has been widely used in object detection. 
In existing works \cite{zeiler2013visualizing, Krizhevsky:ImageNetCNN, he2014spatial}, the same deep architecture was used. Models were different in cropping images at different  locations or using different learned parameters. In our model averaging scheme, we learn models under multiple settings. The settings of the models used for model averaging are shown in \cite{DeepIDNet_web}. They are different in net structures, pretraining schemes, loss functions for the deep model training, adding def-pooling layer or not, and doing bounding box rejection or not. 
Models generated in this way have higher diversity and are complementary to each other in improving the detection results. 


%

The 4 models are automatically selected by greedy search on ImageNet Det val$_2$, and the mAP of model averaging is $50.3\%$ on the test data of  ILSVRC2014, while the mAP of the best single model is $48.2\%$.

\section{Experimental results}
\emph{Overall result on PASCAL VOC.} For the VOC-2007 detection dataset, we follow the approach in \cite{girshick2014rich} for splitting the training and testing data. Table \ref{Table:allMethodVOC07} shows the experimental results on VOC-2007 testing data, which include approaches using hand-crafted features \cite{Girshick:DPMweb, ren2013histograms, wang2013regionlets, vandeSandeCVPR2014, LatSVMObj}, deep CNN features \cite{girshick2014rich, he2014spatial}, and CNN features with deformation learning \cite{girshick2014deformable}. Since all the state-of-the-art works reported single-model results on this dataset, we also report the single-model result only. Our model was pretrained on bounding box annotation, with deformation, without context, and with ZF as the baseline net. Ours outperforms RCNN \cite{girshick2014rich} and SPP \cite{he2014spatial} by about 5\% in mAP. RCNN, SPN and our model are all pre-trained on the ImageNet Cls-Loc training data and fine-tuned on the VOC-2007 training data.

\begin{table*}
\setlength{\tabcolsep}{1.5pt}
\centering
\caption{Detection mAP ($\%$) on ILSVRC2014  for top ranked approaches with single model (sgl) and average model (avg). 
}
{\small
\begin{tabular}{*7c|cc}
\hline	 
	     approach  & Flair \cite{vandeSandeCVPR2014} & RCNN\cite{girshick2014rich}  &Berkeley Vision&UvA-Euvision	& DeepInsight&GoogLeNet\cite{szegedy2014going} &ours& \\
\hline   ImageNet val$_2$ (avg)    &n/a & n/a &n/a&n/a & 42&44.5  & 50.7\\
\hline   ImageNet val$_2$  (sgl)   &n/a & 31.0 &33.4&n/a & 40.1&38.8  & 48.2\\
\hline   ImageNet test  (avg)     &22.6   & n/a  &n/a&n/a& 40.5 &43.9  & 50.3\\
\hline   ImageNet test  (sgl)     &n/a   & 31.4  &34.5&35.4& 40.2 &38.0  & 47.9\\
\hline
\end{tabular}
}
\label{Table:allMethod}
\end{table*}

\begin{table*}
\setlength{\tabcolsep}{1.5pt}
\centering
\caption{Detection mAP ($\%$) on PASCAL VOC-2007 test set. }
{\small
\begin{tabular}{*7c|c}
\hline	 HOG-DPM \cite{Girshick:DPMweb}  & HSC-DPM \cite{ren2013histograms} & Regionlet \cite{wang2013regionlets} &Flair \cite{vandeSandeCVPR2014} &DP-DPM \cite{girshick2014deformable}&	RCNN\cite{girshick2014rich} & SPP \cite{he2014spatial} &ours (single model)\\
\hline   33.7                     & 34.3   & 41.7 & 33.3&45.2                   &58.5                          & 59.2 & 64.1\\
\hline
\end{tabular}
}
\label{Table:allMethodVOC07}
\vspace{-10pt}
\end{table*}

\emph{Experimental Setup on ImageNet.}
The ImageNet Large Scale Visual Recognition Challenge (ILSVRC) 2014 \cite{ILSVRCarxiv14} contains two different datasets: 1) the classification and localization (Cls-Loc) dataset and 2) the detection (Det) dataset. 
The training data of Cls-Loc contains 1.2 million images with labels of $1,000$ categories. It is used to pretrain deep models. The same split of train and validation data from the Cls-Loc is used for image-level annotation and object-level annotation pretraining.
The Det contains 200 object categories and is split into three subsets, train, validation (val), and test data. We follow  RCNN \cite{girshick2014rich} in splitting the val data into val$_1$ and val$_2$. Val$_1$ is used to train models, val$_2$ is used to evaluate separate components, and test is used to evaluating the overall performance. The val$_1$/val$_2$ split is the same as that in \cite{girshick2014rich}.


\emph{Overall result on ImageNet Det.} RCNN \cite{girshick2014rich} is used as the state-of-the-art for comparison. The source code provided by the authors was used to and we were able to repeat their results. 
Table \ref{Table:allMethod} summarizes the results from ILSVRC2014 object detection challenge. It includes the best results on the test data submitted to ILSVRC2014 from GoogLeNet \cite{szegedy2014going}, DeepInsignt, UvA-Euvision, and Berkeley Vision, which ranked top among all the teams participating in the challenge. In terms of single-model and model averaging performance, we achieve the highest mAP. It outperforms the winner of ILSVRC2014, GoogleNet, by 6.1\% on mAP.  

\begin{table}
\setlength{\tabcolsep}{1.5pt}
\centering
\caption{Study of bounding box (bbox) rejection and baseline deep model on ILSVRC2014 val$_2$ without context or def-pooling. }
{\small
\begin{tabular}{*6c}
\hline	 bbox rejection? & n & y& y & y& y\\
				 deep model  & A-net & A-net& Z-net&O-net&G-net \\
\hline   mAP ($\%$)        & 29.9  & 30.9 & 31.8 &36.6&37.8\\
         meadian AP ($\%$)  & 28.9  & 29.4 & 30.5 &36.7&37 \\
\hline
\end{tabular}
}
\label{Table:RejAndModel}
\vspace{-10pt}
\end{table}

\begin{table*}
\setlength{\tabcolsep}{1.5pt}
\centering
\caption{Ablation study of the two pretraining schemes in Section \ref{Sec:Prtrain} for different net structures on ILSVRC2014 val$_2$. Scheme 0  only uses image-level annotation for pretraining. Scheme 1 uses object-level annotation for pretraining. Context is not used. }
{\small
\begin{tabular}{*9ccccc}
\hline   net structure  	& A-net & A-net & A-net &Z-net&Z-net& Z-net & Z-net  &O-net&O-net&G-net&G-net \\
			   class number 		& 1000 & 1000  & 1000  &1000 & 200& 1000 & 1000&1000&1000&1000&1000\\
				 bbox rejection   &n     &n      &y      &y    &n   &n &y  &y &y&y &y\\
				 pretrain scheme  & 0    &  1    &1      &0    &1   &1 &1  &0 &1&0 &1\\
\hline   mAP ($\%$) 			&29.9	 & 34.3  &34.9	 &31.8 &29.9&35.6	& 36.0 &36.6&39.1&37.8&40.4\\
			   meadian AP ($\%$)&28.9	 & 33.4  &34.4	 &30.5 &29.7&34.0		& 34.9 &36.7&37.9&37.0&39.3 \\
\hline
\end{tabular}
}
\label{Table:PretrainScheme}
\vspace{-10pt}
\end{table*}

\begin{table}
\setlength{\tabcolsep}{1.5pt}
\centering
\caption{Investigation on baseline net structures with dep-pooling on ILSVRC2014 val$_2$. Use pretraining scheme 1 but no context.}
{\small
\begin{tabular}{*7c}
\hline   net structure  	& Z-net&   D-Def(Z) &O-net &D-Def(O)&G-net &D-Def(G)\\
\hline   mAP ($\%$) 						  & 36.0&  38.5  &39.1& 41.4&40.4& 42.7 \\
			   meadian ($\%$) 			 & 34.9 &  37.4&37.9& 41.9&39.3& 42.3\\
\hline
\end{tabular}
}
\label{Table:designs}
\vspace{-10pt}
\end{table}

\begin{table*}
\setlength{\tabcolsep}{1.5pt}
\centering
\caption{Ablation study of the overall pipeline for single model on ILSVRC2014 val2. It shows the mean AP after adding each key component step-by-step.}
{\small
\begin{tabular}{*8ccccc}
\hline   detection pipeline  	& RCNN& +bbox     & A-net & Z-net& O-net& image to bbox  &+edgbox& +Def  &+multi-scale& +context & +bbox   \\
															&     & rejection & to Z-net& to O-net& to G-net & pretrain &candidate&  pooling &pretrain&  & regression \\
\hline   mAP ($\%$) 					    	&29.9& 30.9& 31.8& 36.6& 37.8& 40.4 & 42.7&44.9 &47.3&47.8&48.2\\
			   meadian AP ($\%$) 			    &28.9& 29.4& 30.5& 36.7& 37.0& 39.3 & 42.3&45.2 &47.8&48.1&49.8\\
\hline   mAP improvement ($\%$) 		&   & +1&+0.9& +4.8& +1.2 & +2.6 &+2.3&+2.2&+2.4&+0.5&+0.4\\
\hline
\end{tabular}
}
\label{Table:overall}
\vspace{-10pt}
\end{table*}

\subsection{Ablation study}
The ImageNet Det is used for ablation study. Bounding box regression is not used if not specified.

\subsubsection{Baseline deep model and bounding box rejection}
As shown in Fig. \ref{Fig:DeepIDmodel}, a baseline deep model is used in our DeepID-Net. 
Table \ref{Table:RejAndModel} shows the results for different baseline deep models and bounding box rejection choices. 
AlexNet in \cite{Krizhevsky:ImageNetCNN} is denoted as A-net, ZF in \cite{zeiler2013visualizing} is denoted as Z-net, and overfeat in \cite{sermanet2013overfeat} is denoted as O-net.
Except for the two components investigated in Table \ref{Table:RejAndModel}, other components are the same as RCNN, while the new training schemes and the new components introduced in Section \ref{Sec:DeepIdNet} are not included. The configuration in the second column of Table \ref{Table:RejAndModel} is the same as RCNN (mean mAP $29.9\%$).
Based on RCNN, applying bounding box rejection improves mAP by 1\%. Therefore, bounding box rejection not only saves the time for training and validating new models, which is critical for future research, but also improves detection accuracy.  Both with bounding box rejection, ZF \cite{zeiler2013visualizing} performs better than AlexNet \cite{Krizhevsky:ImageNetCNN}, with 0.9\% mAP improvement. Overfeat \cite{sermanet2013overfeat} performs better than ZF, with 4.8\% mAP improvement. GoogleNet \cite{szegedy2014going} performs better than Overfeat, with 1.2\% mAP improvement.



\subsubsection{Investigation on different pretraining schemes and baseline net structures}
There are two different sets of data used for pretraining the baseline deep model: 1) the ImageNet Cls train data with 1000 classes and 2) the ImageNet Cls train data data with the same 200 classes as Det. There are two different annotation levels, image and object. Table \ref{Table:PretrainScheme} show the results for investigation on image class number, annotation levels, and net structures. When producing these results, other new components introduced in Section \ref{Sec:DeformLayer}-\ref{Sec:Context} are not included. For pretraining, we drop the learning rate by 10 when the classification accuracy of validation data reaches plateau, until no improvment is found on the validation data. For fine-tuning, we use the same initial learning rate (0.001) and the same number of iterations (20,000) for dropping the learning rate by 10 for all net structures, which is the same setting in RCNN \cite{girshick2014rich}. 


Using object-level annotation, pretraining on 1000 classes performs better than pretraining on 200 classes by 5.7\% mAP. Using the same 1000 classes, pretraining on object-level-annotation performs better than pretraining on image-level annotation by 4.4\% mAP for A-net and 4.2\% for Z-net.
This experiment shows that object-level annotation is better than image-level annotation in pretraining deep model. Pretraining with more classes improves the generalization capability of the learned feature representations.


%

\subsubsection{Investigation on def-pooling layer}
Different deep model structures are investigated and results are shown in Table \ref{Table:designs} using the new pretraining scheme in Section \ref{Sec:Prtrain}. Our DeepID-Net that uses def-pooling layers as shown in Fig. \ref{Fig:DeepIDmodel} is denoted as D-Def. 
Using the Z-net as baseline deep moel, the DeepID-Net that uses def-pooling layer in Fig. \ref{Fig:DeepIDmodel} improves mAP by 2.5\%. Def-pooling layer improves mAP by 2.3\% for both O-net and G-net. This experiment shows the effectiveness of the def-pooling layer for generic object detection. 

\subsubsection{Investigation on the overall pipeline}
Table \ref{Table:overall} summarizes how performance gets improved by adding each component step-by-step into our pipeline. RCNN has mAP $29.9\%$. With bounding box rejection, mAP is improved by about $1\%$, denoted by $+1\%$ in Table \ref{Table:overall}. Based on that, changing A-net to Z-net improves mAP by $0.9\%$. Changing Z-net to O-net improves mAP by $4.8\%$. O-net to G-net improves mAP by $1.2\%$. Replacing image-level annotation by object-level annotation in pretraining, mAP is increased by $2.6\%$. By combining candidates from selective search and edgeboxes \cite{ZitnickDollarECCV14edgeBoxes}, mAP is increased by $2.3\%$. The def-pooling layer further improves mAP by $2.2\%$. Pretraining the object-level annotation with multiple scales 
\cite{chatfield2014return} improves mAP by $2.2\%$.  After adding the contextual information from image classification scores, mAP is increased by $0.5\%$. Bounding box regression improves mAP by $0.4\%$. With model averaging, the final result is $50.7\%$.


%

\section{Appedix A: Relationship between the deformation layer and the DPM}
The quadratic deformation constraint in \cite{LatSVMObj} can be represented as follows:
{\small
\begin{equation}
\label{eq:DefMap5}
\begin{split}
\tilde{m}^{(i,j)} \!=\! m^{(i,j)}-a_{1}(i\!-\!b_{1}\!+\!\frac{a_{3}}{2a_{1}})^2\!-\! a_{2} (j\!-\!b_{2}\!+\!\frac{a_{4}}{2a_{2}})^2,
\end{split}
\vspace{-10pt}
\end{equation}}
\!\!where $m^{(i,j)}$ is the $(i,j)$th element of the part detection map $\mathbf{M}$, $(b_{1}, b_{2})$ is the predefined anchor location of the $p$th part. They are adjusted by $a_3/2a_1$ and $a_4/2a_2$, which are automatically learned.
$a_{1}$ and $a_{2}$ (\ref{eq:DefMap5}) decide the deformation cost. There is no deformation cost if $a_{1}=a_{2}=0$. Parts are not allowed to move if $a_{1}=a_{2}=\infty$. $(b_{1}, b_{2})$ and $(\frac{a_{3}}{2a_{1}}, \frac{a_{4}}{2a_{2}})$ jointly decide the center of the part. 
The quadratic constraint in Eq. (\ref{eq:DefMap5}) can be represented using Eq. (\ref{eq:GenDefMap2}) as follows:
{\small
\begin{align}
\tilde{m}^{(i,j)} \! &=\! m^{(i,j)}-a_{1}d_{1}^{(i,j)}- a_{2}d_{2}^{(i,j)}- a_{3}d_{3}^{(i,j)}  \!- a_{4} d_{4}^{(i,j)} \! -\! a_5, \nonumber \\
d_{1}^{(i,j)}\!  &=\! (i-b_{1})^2, \ \ d_{2}^{(i,j)} \!= \!(j-b_{2})^2, d_{3}^{(i,j)}\!  =\! i-b_{1}, \nonumber\\
d_{4}^{(i,j)} &\!= \!j-b_{2}, a_5={a_3}^2/(4a_1)+{a_4}^2/(4a_2).
\label{eq:DefMap6}
\vspace{-10pt}
\end{align}}
 In this case, $a_1, a_2, a_3$ and $a_4$ are parameters to be learned and $d_{n}^{(i,j)}$ for $n=1, 2, 3, 4$ are predefined. $a_5$ is the same in all locations and need not be learned. 
The final output is:

\vspace{-10pt}
\begin{equation}
\label{eq:DefMap3}
b = \max_{(i,j)} {\tilde{m}^{(i,j)}},
\vspace{-5pt}
\end{equation}
where $\tilde{m}^{(i,j)}$ is the $(i, j)$th element of the matrix $\tilde{\mathbf{M}}$ in (\ref{eq:DefMap5}).


\section{Conclusion}
This paper proposes a deep learning based object detection pipeline, which integrates the key components of bounding box reject, pretraining, deformation handling, context modeling, bounding box regression and model averaging. It significantly advances the state-of-the-art from mAP $31.0\%$ (obtained by RCNN) to $50.3\%$ on the ImgeNet object task. Its single model and model averaging performances are the best in ILSVC2014. A global view and detailed component-wise experimental analysis under the same setting are provided to help  researchers understand the pipeline of deep learning based object detection. 

We enrich the deep model by introducing the def-pooling layer, which has great flexibility to incorporate various deformation handling approaches and deep architectures. 
Motivated by our insights on how to learn feature representations more suitable for the object detection task and with good generalization capability, a pretraining scheme is proposed. 
By changing the configurations of the proposed detection pipeline, multiple detectors with large diversity are obtained, which leads to more effective model averaging.



\textbf{Acknowledgment}: This work is supported by the General Research Fund sponsored by the Research Grants Council of Hong Kong (Project Nos.  CUHK14206114 and CUHK14207814).

{\small
\bibliographystyle{ieee}
\bibliography{./PME}
}

\end{document}